\documentclass[review]{elsarticle}

\usepackage[T1]{fontenc}    
\usepackage{hyperref}       
\usepackage{url}            
\usepackage{booktabs}       
\usepackage{amsfonts}       
\usepackage{nicefrac}       
\usepackage{microtype}      
\usepackage{graphicx}
\usepackage{subfigure}
\usepackage{amsmath}
\usepackage{multirow}
\usepackage{tabularx}
\usepackage{array}

\journal{Journal of \LaTeX\ Templates}

\begin{document}

\begin{frontmatter}

\title{SymmetricNet: A mesoscale eddy detection method based on multivariate fusion data}
\author[mymainaddress]{Zhenlin Fan}
\author[mymainaddress]{Guoqiang Zhong\corref{mycorrespondingauthor}}
\cortext[mycorrespondingauthor]{Corresponding author}
\ead{gqzhong@ouc.edu.cn}




\address[mymainaddress]{Department of Computer Science and Technology, Ocean University of China,238 Songling Road, Qingdao 266100, China}

\begin{abstract}
Mesoscale eddies play a significant role in marine energy transport, marine biological environment and marine climate. Due to their huge impact on the ocean, mesoscale eddy detection has become a hot research area in recent years. Therefore, more and more people are entering the field of mesoscale eddy detection. However, the existing detection methods mainly based on traditional detection methods typically only use Sea Surface Height (SSH) as a variable to detect, resulting in inaccurate performance. In this paper, we propose a mesoscale eddy detection method based on multivariate fusion data to solve this problem. We not only use the SSH variable, but also add the two variables: Sea Surface Temperature (SST) and velocity of flow, achieving a multivariate information fusion input. We design a novel symmetric network, which merges low-level feature maps from the downsampling pathway and high-level feature maps from the upsampling pathway by lateral connection. In addition, we apply dilated convolutions to network structure to increase the receptive field and obtain more contextual information in the case of constant parameter. In the end, we demonstrate the effectiveness of our method on dataset provided by us, achieving the test set performance of 97.06\% , greatly improved the performance of previous methods of mesoscale eddy detection.
\end{abstract}

\begin{keyword}
Deep leaning, Mesoscale eddy detection, Multivariate fusion data
\end{keyword}

\end{frontmatter}


\section{Introduction}

With the development of deep learning~\cite{lecun2015deep}, deep convolutional neural networks (DCNNs)~\cite{lecun1998gradient} become an increasingly popular method of artificial intelligence in recent years. Moreover, deep learning has been widely used in many practical problems such as pattern recognition and computer vision, and has achieved excellent results. Among others, semantic segmentation as an important branch of computer vision, has benefited from the rapid development of deep learning~\cite{everingham2015pascal}. The semantic segmentation method is specific to the pixel level when processing an image, in other words, the method assigns each pixel in the image to an object class~\cite{mottaghi2014role,cordts2016cityscapes,caesar2018coco}. Before deep learning was applied to the field of computer vision, people generally used Texton Forest~\cite{shotton2008semantic} or Random Forest~\cite{schroff2008object} methods to construct classifiers for semantic segmentation. Since full convolutional network~\cite{long2015fully} achieved the state-of-the-art performance of semantic segmentation at that time, a variety of approaches based on deep learning are beginning to make semantic segmentation results better and better~\cite{ronneberger2015u,chen2014semantic,he2017mask}. Mesoscale eddy detection can be regarded as a semantic segmentation problem, because the bounding box in the object detection is a rectangular shape, and it is difficult to accurately detect the irregular shape of the mesoscale eddies.

Mesoscale eddies (also known as weather-type ocean eddies) refer to eddies with a diameter of 100-300 km in the ocean and a life span of 2-10 months~\cite{wyrtki1976eddy,chelton2007global}. They are usually divided into two types: cyclonic eddies (counterclockwise rotation in the northern hemisphere) and anti-cyclonic eddies (counterclockwise rotation in the southern hemisphere). Thus, mesoscale eddy detection can be seen as a semantic segmentation problem with only two classes, except background. Currently, mesoscale eddy detection methods based on deep learning are very few, there are few reasonable and effective models for people to detect mesoscale eddy. In recent years, there are relatively representative papers on mesoscale eddy detection, including EddyNet~\cite{lguensat2018eddynet}, DeepEddy~\cite{du2019deep} and pyramid scene parsing network (PSPNet)~\cite{xu2019oceanic}. However, there are also some disadvantages on these network architecture. Both EddyNet and PSPNet regard mesoscale eddy detection as a semantic segmentation problem. But these two methods only use SSH as a detection variable, while there are other influencing factors that will assist in mesoscale eddy detection. Therefore, the accuracy of mesoscale eddy detection is difficult to guarantee for using only one variable for detection. Different from these two methods, DeepEddy uses more than one variable for detection. However, this method that can only perform simple classification is impossible to detect multiple mesoscale eddies in a sea area.

\begin{figure}[ht]
\centering
\subfigure[SSH visualization]{
\includegraphics[width=3cm,height=3cm]{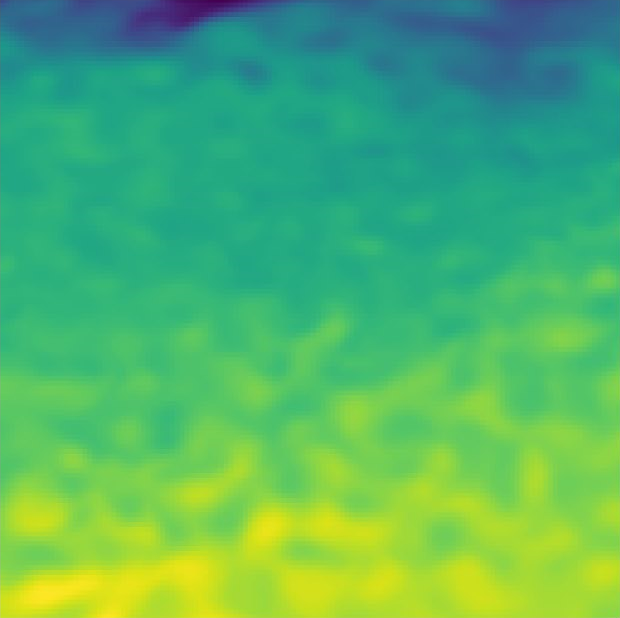}}
\subfigure[Groundtruth]{
\includegraphics[width=3cm,height=3cm]{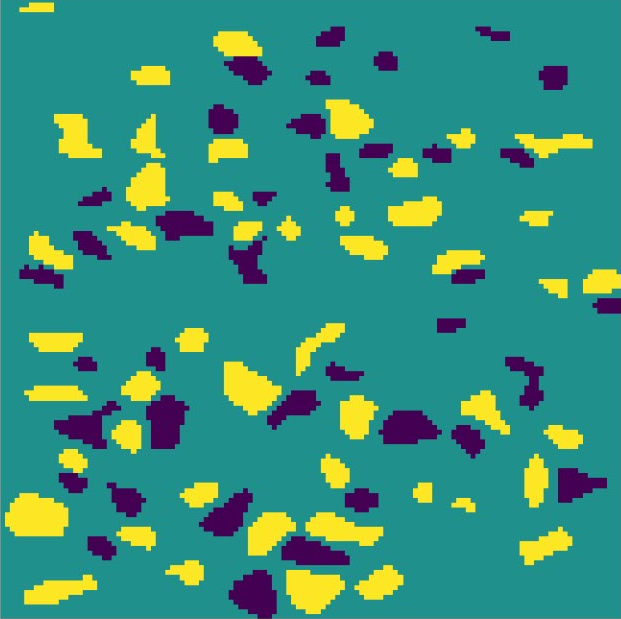}}
 \caption{The visualization of SSH and groundtruth in a certain area of the sea.}\label{fig1}
\end{figure}

In our work, we design a network structure, which input is multivariate fusion data, ultimately geting higher accuracy than the previous methods. One of the main difficulties in mesoscale eddy detection is that there are very few existing datasets to use, and it takes a certain amount of time and effort to collect and label the data. So we build a multivariate fusion dataset including multiple variables to overcome this difficulty. We download the satellite remote sensing data from the Copernicus Marine Environment Monitoring Service (CMEMS), including three variables: SSH~\cite{fu2010eddy,mason2014new}, SST~\cite{voorhis1976influence,dong2011automated} and velocity of flow, which can assist in mesoscale eddy detection. It should be noted that the data of velocity of flow contains two directions, namely the velocity of the eastward seawater and the velocity of the northward seawater. This can be understood that the velocity vector of a certain point in the ocean is decomposed into the east, west, and north, south directions. Thus, different from SSH and SST variables with only one channel respectively, the velocity of flow variable has two channels. One point to mention is that if the data from the fusion of these three variables is visualized, we are worried that the information will be blurred for our vision. Additionally, most of the current methods only use SSH variable for detection due to it's great impact on mesoscale eddy detection. Thus the visual image based on SSH variable are labeled by experts as groundtruth to make it convenient to compare with the those methods. The visual images of SSH and groundtruth in a certain area of the sea are as illustrated in Figure~\ref{fig1}. Figure~\ref{fig1} (a) shows the visual image based SSH variable, Figure~\ref{fig1} (b) shows groundtruth segmentation, where the yellow area represents the anti-cyclonic eddies, the dark blue area represents the cyclonic eddies, and the blue area represents the area without eddies. The network architecture we designed is a symmetric network, which combine low-resolution, semantically strong features with high-resolution, semantically weak features via lateral connections. The benefit of combination is that the final feature map of our model not only has rich semantics, but does not lose a lot of location information. Accordingly, our framework improve detection accuracy significantly. Furthermore, considering that convolution will reduce resolution so that the image is too abstract to be discernible, dilated convolutions~\cite{yu2015multi} are added to upsampling process to increase the receptive field. In contrast to EddyNet and PSPNet, our method combine multiple variables to improve detection accuracy. What is more, the mesoscale eddy detection method based on PSPNet has no innovation for the network, but directly applies the PSPNet~\cite{zhao2017pyramid} proposed by others for semantic segmentation to the mesoscale eddy detection. In contrast to DeepEddy, we can achieve detection in a sea area with multiple mesoscale eddies, including cyclonic eddies and anti-cyclonic eddies.

In a word, the main contributions of our work are:
\begin{itemize}
\item We build a multivariate fusion dataset based on multiple variables, provided annotation masks for the visual image of SSH by experts.
\end{itemize}

\begin{itemize}
\item We propose a novel symmetric network , which merges low-level feature maps from the downsampling pathway and high-level feature maps from the upsampling pathway by lateral connection. Besides that, we also apply dilated convolutions to network architecture to increase the receptive field and obtain more contextual information in the case of constant parameter.
\end{itemize}

\begin{itemize}
\item Our approach outperforms all previous methods, attain state-of-art performance on multivariate fusion dataset.
\end{itemize}

\section{Related work}

In this section, we introduce the development of mesoscale eddy detection in detail. Due to the negligible effect of mesoscale eddies on human survival and development, the detection of mesoscale eddies has been widely studied for many years. In the early days, traditional methods were used to detect mesoscale eddies. So its accuracy needs to be improved because of the limitations of the method. With the rapid development of deep learning, more and more mesoscale eddy detection based on deep learning methods has begun to appear.

\subsection{Mesoscale eddy detection based on traditional method}

In the early stage, mesoscale eddy detection based ocean remote sensing images mainly relied on expert visual interpretation methods. This method is not only labor intensive, but also has inevitable human factors. Therefore, people began to use mathematical knowledge to perform mesoscale eddy detection. Nichol~\cite{nichol1987autonomous} used a computer-searched region of the same gray value in the image, and attempted to extract a similar eddy structure from the relationship diagram generated between these regions. Due to the complexity of the imaging process of ocean remote sensing images, it is difficult to extract the eddy detection features based on the connected regions of gray values such as images. So, Peckinpaugh and Holyer~\cite{peckinpaugh1994circle} proposed a method for eddy detection using Hough transform circle detection operator~\cite{illingworth1988survey} based on the edge detection of remote sensing image. Due to the complexity of the eddy shape, the edge detection curve is generally not a regular circle, so the method is relatively rough. In addition, one of the main disadvantages of the Hough transform is that as the amount of processed data increases, the amount of storage and computation required increases dramatically, and the detection error also increases. Next, Ji et al.~\cite{Ji2002detection} proposed a mesoscale eddy detection method of phased refinement based on the analysis of that the edge of the eddy region of the remote sensing image is generally composed of a segmented arc curve of approximate ellipse. Firstly, the candidate sub-region is generated by the method of curve fitting and local region Hough transform, the large-area remote sensing image is reduced to several local regions, then the local sub-regions are selected, and finally the eddy regions are detected by the gray-scale segmentation in the selected sub-regions.

In order to improve the accuracy of detection, people began to use more accurate methods for mesoscale eddy detection. Generally speaking, researchers classify mesoscale eddy detection methods into two categories based on data characteristics, methods based on Euler data and methods based on Lagrangian data. For Euler data of observations or numerical simulations, the eddy is identified from the image of the two-dimensional or three-dimensional field based on the physical characteristics of the flow field. The main methods are Okubo-Weiss parameter value methods~\cite{isern2003identification,penven2005average,chelton2007global}, mesoscale eddy detection using sea surface height variation~\cite{chelton2011global}, wavelet analysis method of relative vorticity~\cite{doglioli2007tracking}, wind angle method of geometric or kinematics of the flow field~\cite{chaigneau2008mesoscale}, edge detection~\cite{canny1986computational} and so on. For Lagrangian data, according to the physical characteristics of the flow field, there are mainly rotation methods~\cite{griffa2008cyclonic}, Lagrangian stochastic model method~\cite{lankhorst2006self}, spiral trajectory search method based on geometric features of trajectory~\cite{dong2011scheme} and so on. On this basis, Liu detected eddy by the geometric features of the eddy in the vector field based on Lagrangian data and Euler data. In the same year, Faghmous et al.~\cite{faghmous2012eddyscan} proposed a method named EddyScan, a physically consistent ocean eddy monitoring application. They tracked eddies globally as closed contour of SSH anomalies. This was done in two steps: first they identified features that displayed the spatial properties of an ocean mesoscale eddy. This was accomplished by assigning binary values to the SSH data based on whether or not a varying threshold was exceeded, and subsequently identifying mesoscale connected component features. They then pruned the identified connected components based on other criteria that are physically consistent with eddies at a given latitude. This method is already quite good in the traditional mesoscale eddy detection method.

Overall, mesoscale eddy detection based on traditional method has some defects not only from the selection of the detection variables but also the detection methods, which need to be improved by people's continuous efforts.

\subsection{Mesoscale eddy detection based on deep learning method}

Although traditional mesoscale eddy detection algorithms have achieved better performance than before, the widespread use of deep learning clearly provides a more efficient method for mesoscale eddy detection. Consequently, as expected, mesoscale eddy detection based on deep learning began to emerge.

So far, there are not many mesoscale eddy detection methods based on deep learning. In recent years, the representative one is the EddyNet proposed by Lguensat et al.~\cite{lguensat2018eddynet}, a deep learning based architecture for automated eddy detection and classification from SSH maps provided by the CMEMS. This method is actually a simple network architecture based on U-Net. Although there is no much innovation in network structure, this method combines the convolutional network in deep learning with mesoscale eddy detection and uses it as a semantic segmentation problem early, which greatly improves the accuracy and speed of mesoscale eddy detection compared with the previously complicated traditional methods. Subsequently, Franz et al.~\cite{franz2018ocean} developed an eddy identification and tracking framework with two different approaches that are mainly based on feature learning with CNNs. One approach is the CNN-module, which can learn a model from which they can detect eddy cores at a single epoch. Another approach is the Kanade-Lucas-Tomasi (KLT)~\cite{lucas1981iterative} -tracker, which can track the detected eddies using a sparse optical flow. In other words, a recurrent neural network (RNN) uses convolutional long short-term memory (LSTM) units to track the eddy cores. Both methods have achieved good results, but a common disadvantage of these two methods is that the input variable are only one. The former variable is SSH, and the latter variable is global sea level anomaly (SLA) maps. Beside that, there are more than one factor that can identify mesoscale eddies, so using only one variable for mesoscale eddy detection will inevitably reduce the accuracy of detection. To solve this problem, Du et al.~\cite{du2019deep} proposed a DeepEddy method with multi-scale feature fusion in remote sensing for automatic oceanic eddy detection, which is an architecture based on a simple principal component analysis network (PCANet)~\cite{chan2015pcanet}. Although the variables of this method combine SSH and SST and is no longer just a single variable, unfortunately this method can only perform one classification task. Recently, Xu et al.~\cite{xu2019oceanic}adapted PSPNet to mesoscale eddy detection, which is an architecture for semantic segmentation. There are two disadvantages about this method: input variable is only one same as before and there are not much innovation on network. But this method followes EddyNet and is another one considering mesoscale eddy detection as a segmentation problem.

In the consequence, there is no approach to combine multivariate fusion input and mesoscale eddy detection. Our approach combines deep learning with fused data and have some innovation in the network architecture, ultimately achieving the state-of-art results on our dataset consisting of multiple variables.


\section{Method}

\begin{figure}[ht]
\centering
\includegraphics[width=13cm,height=7cm]{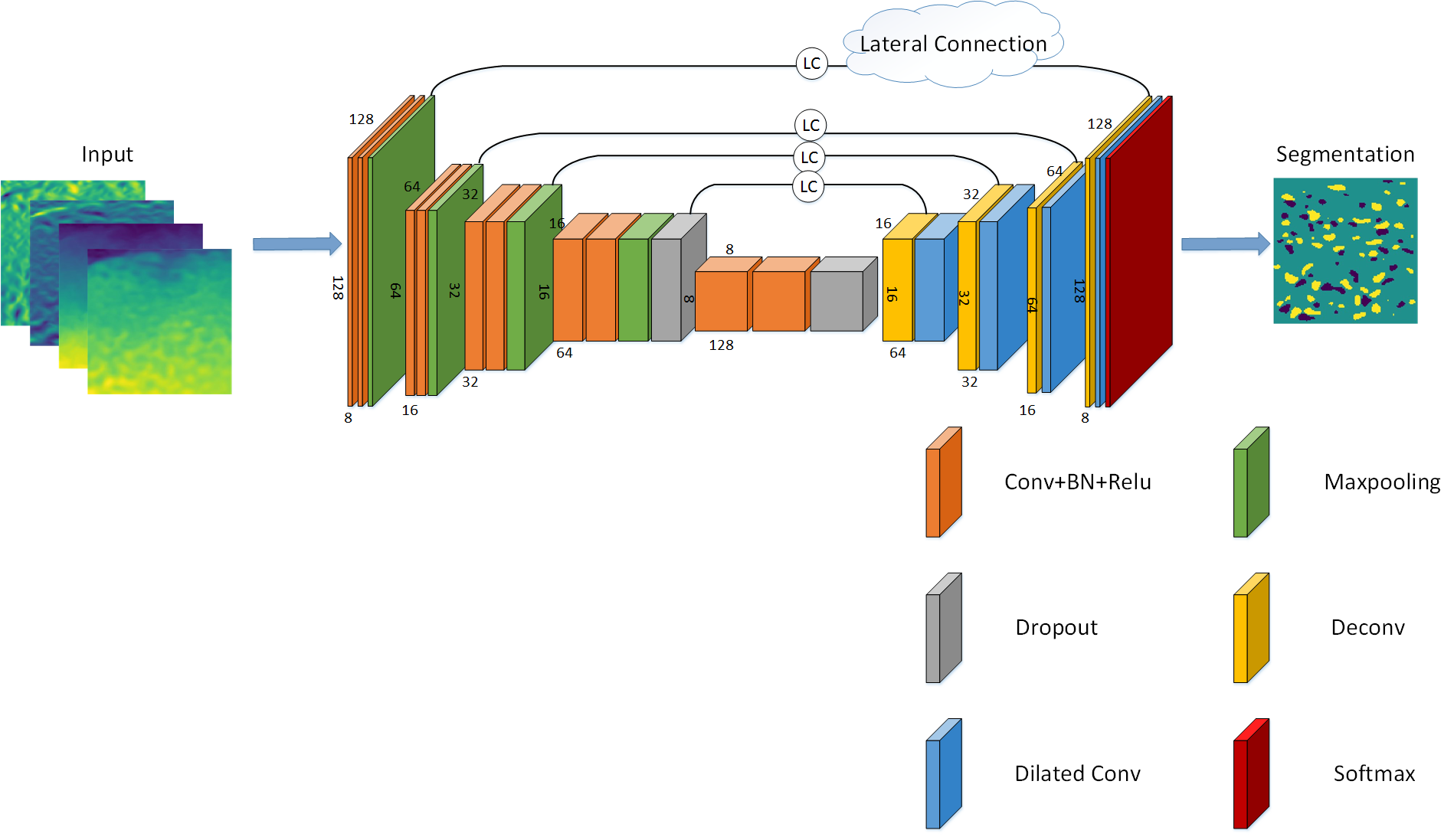}
\caption{The proposed network architecture.}
\label{fig2}
\end{figure}

In this section, we introduce our method in detail. The method we propose is a symmetric network, is shown in Figure~\ref{fig2}. We combine the features of multiple variables as input and obtain a segmentation with higher accuracy through our network. Meanwhile, in order to get a more accurate segmentation, we merge the cross entropy loss function and the dice loss function as the loss function of our method, and finally demonstrate the effect of loss function by comparative experiment.

\subsection{Network architecture}

\subsubsection{Symmetric network architecture}

Our proposed network architecture is a symmetric network. Our network is composed of the downsampling pathway (left side) and the upsampling pathway (right side) and a transition block (in the middle). In short, it is similar to the encoder-decoder network. In fact, the downsampling pathway is a series of convolutions network. It contains 4 downsampling blocks, of which the convolutional operation mainly consists of two $3\times3$ convolutions, each followed by a batch normalization (BN) layer and a rectified linear unit (ReLU). One of the things worth mentioning is the BN layer. The main role of BN layer is to alleviate the gradient disappearance/explosion phenomenon in network training and speed up the training speed of the model. Next, a $2\times2$ max pooling operation with stride 2 as the following layer for downsampling. What't more, in order to avoid over-fitting in our network, the fourth downsampling block is applied to a dropout layer, which becomes the last layer. So the last layer in other downsampling blocks is still the max pooling layer. Additionally, the number of channels is doubled when performing a downsampling block. Note that, the downsampling pathway can be summarized as that the length and width of feature map become half of the original and the number of channels is doubled when passing a downsampling block. In a similar way, the inverse of the downsampling pathway is actually the upsampling pathway. So, upsampling pathway contains 4 upsampling blocks, each of which consists of a deconvolutional operation that doubles the length and width of feature map and halves the number of channels, a lateral connection with the feature maps from the downsampling pathway, and a $3\times3$ dilated convolution with a rate of 4. Similarly, the upsampling pathway can be summarized as that the length and width of feature map is doubled, and the number of channels becomes half of the original when passing a upsampling block. Aside from these four downsampling blocks and four upsampling blocks, there is a transition block following the fourth downsampling block, which consists of two $3\times3$ convolutions, each followed by a BN layer and a ReLU layer. Same as the fourth downsampling block, there is a dropout layer at the end of the transition block to avoid over-fitting in our network.

In the end, we take the output of the last upsampling block as input and put it into the final softmax layer to achieve pixel-level classification, finally attain a segmentation map.

\subsubsection{Lateral connection}

\begin{figure}[ht]
\centering
\includegraphics[width=13cm,height=7cm]{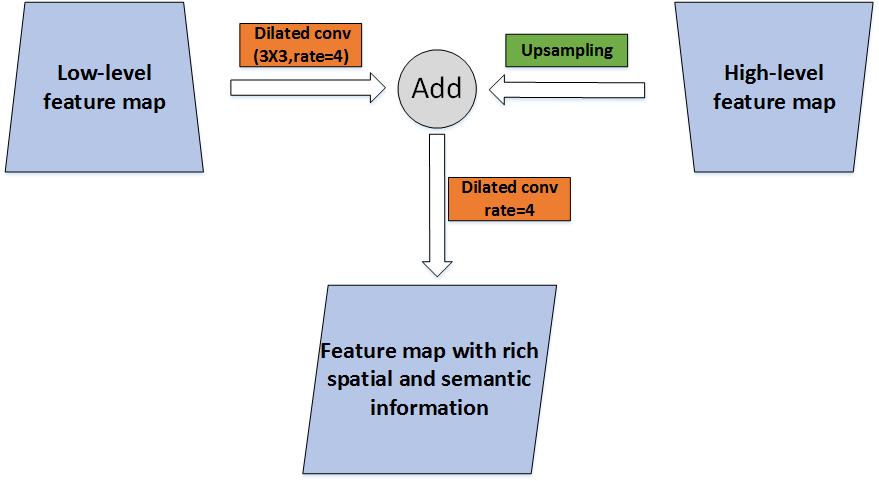}
\caption{The lateral connection.}
\label{fig3}
\end{figure}

There are some unavoidable problems in the upsampling process and the downsampling process. We find that the convolutional operation becomes more and more popular in deep learning because it can effectively extract rich semantic information of the image. However, with the image continuously downsampling, the semantic information is getting richer and richer, and the resolution is getting smaller and smaller. So it is possible to lose some of the more important spatial information. Similarly, although the resolution is improved when the upsampling is performed, some important details may be lost during deconvolution. After the above analysis, we realize that low-level feature maps can get the segmentation results of the larger targets and high-level feature maps can get the segmentation results of the smaller targets. Therefore, we raise the idea of merging the low-level images and high-level images. We combine their
strengths to eliminate their shortcomings. Figure~\ref{fig3} shows a lateral connection of the low-level feature maps from the downsampling pathway and high-level feature maps from upsampling pathway in detail.

Firstly, the high-level images output from the transition block or upsampling blocks are doubled the length and width and halved the number of channels by a upsampling operation. We find the corresponding feature maps in the downsampling pathway according to it's size because the size of feature maps that need to be added must be the same. Then, we apply $3\times3$ dilated convolution with a rate of 4 to low-level feature maps with rich spitial information, performing semantic extraction slightly without changing the resolution to strive to mitigate the disadvantages of low-level feature maps. Subsequently, low-level feature maps are added the high-level feature maps with rich semantic information of the same size obtained by upsampling element by element. Last but not least, the merged feature maps undergo a $3\times3$ dilated convolution with a rate of 4 to increase the receptive field while maintaining the number of convolution kernel parameter.

From here we see that the lateral connection of the low-level feature maps with rich spatial information from the downsampling pathway and high-level feature maps with rich semantic information from the upsampling pathway can combine the advantages of both, geting feature maps with rich spatial and semantic information.

\subsubsection{Dilated convolution}

In the field of deep learning, convolutional operations have become an indispensable part, playing an increasingly important role. In brief, the size of the convolution kernel is generally smaller than the size of the input image, so the feature extracted by the convolution will pay more attention to the local part. In fact, there is no need to perceive the global image for each neuron. It only needs to perceive the local part, and then is combined with other neurons with local information at a high level to obtain global information. Unfortunately, every coin has two sides. When convolutional operation is performed, spatial information is gradually reduced as the semantic information is gradually enriched due to the decline in resolution of the feature map. In response to the above question, we replace normal convolution with dilated convolution to increase the receptive field of the convolution kernel and reduce the loss of spatial information of the feature map.

\begin{figure}[ht]
\centering
\subfigure[Conv($3\times3$)]{
\includegraphics[width=4.5cm,height=4.2cm]{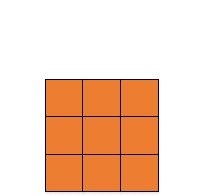}}
\subfigure[Dilated conv($3\times3$,rate=2)]{
\includegraphics[width=6cm,height=5.5cm]{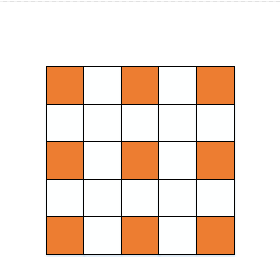}}
 \caption{The comparison of normal convolution and dilated convolution.}\label{fig4}
\end{figure}

The dilated convolution introduces the dilation rate on the basis of linear convolution to increase the receptive field of the convolution kernel, thus obtaining more information on the feature map. As illustrated in Figure~\ref{fig4}, there is the difference between normal convolution and dilated convolution. Figure~\ref{fig4} (a) shows the $3\times3$ convolution kernel of common convolution and Figure~\ref{fig4} (b) shows the $3\times3$ convolution kernel of dialted convolution with a rate of 2. The orange area represents the parameter of the convolution kernel, while white area represents the parameter filled with zero. There is an interval between the parameters of the dilated convolution, which is the dilated rate minus one. There is no doubt that the receptive field of the convolution kernel becomes larger due to the expansion of convolution kernel. Of course, the increase in receptive field means the enrichment in spatial information. In addition, we can see that the parameters of the normal convolution and the dilated convolution are consistent when the convolution kernel size is fixed. That's because the extra parameters are filled with zeros.

Due to the advantages of the dilated convolution, we apply dilated convolutions to the feature maps with rich spatial and semantic information, avoiding massive loss of spatial information as undergoing normal convolutions. We do our best to ensure that the final feature map has the richest semantic and spatial information at the same time. Consequently, the experimental results we acquire will be as good as possible.

\subsection{Our loss functions}

The mesoscale eddy detection problem is a semantic segmentation problem, which is actually a pixel-level classification problem. Thus, we take the dice loss function as part of our loss function, which seems to be a particularly popular loss function when training pixel-segmented neural networks. However, there is a disadvantage in dice loss function. The gradient of the dice loss function mainly depends on the prediction and the groundtruth. If they are too small, the gradient will change sharply, which will make the training difficult. Besides, the mesoscale eddy detection is actually a 3-class problem, including cyclonic eddies, anti-cyclonic eddies and background. Therefore, the cross-entropy loss function is added to our loss function to reduce the training difficulty of the network, which is the most commonly used loss function for multi-classification problems. To sum up, our loss function is a function merging dice loss function with categorical cross-entropy loss function, which is as follows:
$$ L(P,G)=L_{log}(P,G)-log(1-DL(P,G)).  $$

Among that, $ L_{log}(P,G) $ is the cross-entropy loss function. In our work, we assume that there are K labels, and the probability that the $i$-th pixel is predicted to be the $k$-th label is $p_{i,k}$. $g_{i,k}$ represents the probability of grountruth, it is equal to 1 when $i$-th pixel belongs to the $k$-th class and otherwise it is 0. Then the cross-entropy loss function of the multi-classification problem is: $$ L_{log}(P,G)=-\frac{1}{N} \sum_{i=0}^{N-1} \sum_{k=0}^{K-1} {g_{i,k}log p_{i,k}},  $$
where $G$ is groundtrurh, $P$ is prediction and $N$ is the number of pixels. By the way, $K$ in this paper is 3. As the cross-entropy decreases, the accuracy of our mesoscale eddy detection is also higher.

$ DL(P,G) $ is the dice loss function. We introduce the dice coefficient before introducing the dice loss function, which is a similarity measure function used to calculate the similarity of two samples. In our paper, the dice coefficient describes the similarity between the two contours of prediction and groundtruth. We define $P$ as the prediction and $G$ as the groundtruth.  $|P|$ and $|G|$ are denoted the sum of the elements of $P$ and $G$. So the dice coefficient function is as follows:
$$ DC(P,G)=\frac{2|P\bigcap G|}{|P|+|G|},  $$
According to the above formula, we know that the prediction and groundtruth are exactly the same when the dice coefficient is 1, and the segmentation effect is optimal. In contrast, a dice coefficient of 0 refers to a completely mistaken segmentation, prediction and groundtruth do not coincide at all. In other words, the larger the dice coefficient, the better the performance. As a result, in order to achieve the best performance of our network, we minimize the dice loss shown below:
$$ DL(P,G)=1-DC(P,G)=1- \frac{2|P\bigcap G|}{|P|+|G|}.  $$

\section{Experiments}

In Sec. 4.1, we explain the details of building the dataset. In Sec. 4.2, we introduce the details of training our networks. In Sec. 4.3, we confirm that our method is superior to other methods about mesoscale eddy detection from three aspects of our multivariate fusion dataset, network architecture and loss function.

\subsection{Dataset}

So far, there are very few specific datasets for people to use on mesoscale eddy detection. Therefore, it is essential to build a reliable dataset before proceeding. In most works with mesoscale eddy detection, people mainly rely on SSH for detection, lacking effective information of other factors. According to this problem, we build a multivariate fusion dataset, whose variables are composed of three influencing factors, including SSH, SST and velocity of flow.

Firstly, we download the SSH, SST, and velocity of flow for a total of ten years from January 16, 2000 to December 16, 2009 from CMEMS. The dimensions of these three data are $681 \times 1440 \times 120$, where 681 is the dimension of latitude, 1440 is the dimension of longitude, and 120 represents that the data comes from consecutive 120 months. It should be noted that the velocity of flow contains two directions, namely the velocity of the eastward seawater and the velocity of the northward seawater. This can be understood as the velocity vector of a certain point in the ocean is decomposed into the east, west, and north, south directions. Then, we chose 40 months of data for a three-month interval of 120 months in order to make the data more diverse, due to the mesoscale eddy with a life cycle of up to one year. Lastly, we randomly select data from multiple regions resizing $128 \times 128$ in each type of data of each month without repetition, ensuring that the corresponding positions of SSH, SST and velocity of flow are the same. What't more, for those regions where 80\% of the area does not contain mesoscale eddies, we abandon them to improve the generalization ability of the algorithm. However, there are few
 algorithms that can accurately detect the position of the mesoscale eddies currently, so we invite experts who are proficient in marine knowledge to label the images that our variables visualize. On one hand, images that have been merged by multiple variables may appear ambiguous and illegible when visualizing them. On the other hand, it is easy to conduct comparative experiments considering that most of the current methods only use SSH. Therefore, experts only label the images visualized by SSH variables as groundtruth, which is the most widely used and most influential variable in mesoscale eddy detection. At the time of labeling, the cyclonic eddies are annotated as -1, the anti-cyclonic eddies are annotated as 1, and the background are annotated as 0. In the end, we selected 512 samples as training set, 256 samples as testing set.

\subsection{Details of training}

In our network, we take 8, 16, 32, 64 and 128 convolution kernels for the $ 3\times3 $ convolution of each downsampling block and intermediate transition block. In contrast, the number of convolution kernels of the $ 3\times3 $ convolution of each upsampling block is taken as 64, 32, 16, and 8. The dropout in the last two downsampling blocks are 0.3 and 0.5 respectively.
We train our network using Adam optimizer, which has an initial learning rate of $ 1.0\times 10^{-3}$ and a minimum learning rate of $ 1.0\times 10^{-30}$. Additionally, the batchsize is eventually set to 8 and the epoch is set to 50.

\subsection{Comparative experiment}

\subsubsection{Results on different datasets}

\begin{figure}[ht]
\centering
\subfigure[The loss curve]{
\includegraphics[width=5.5cm,height=5cm]{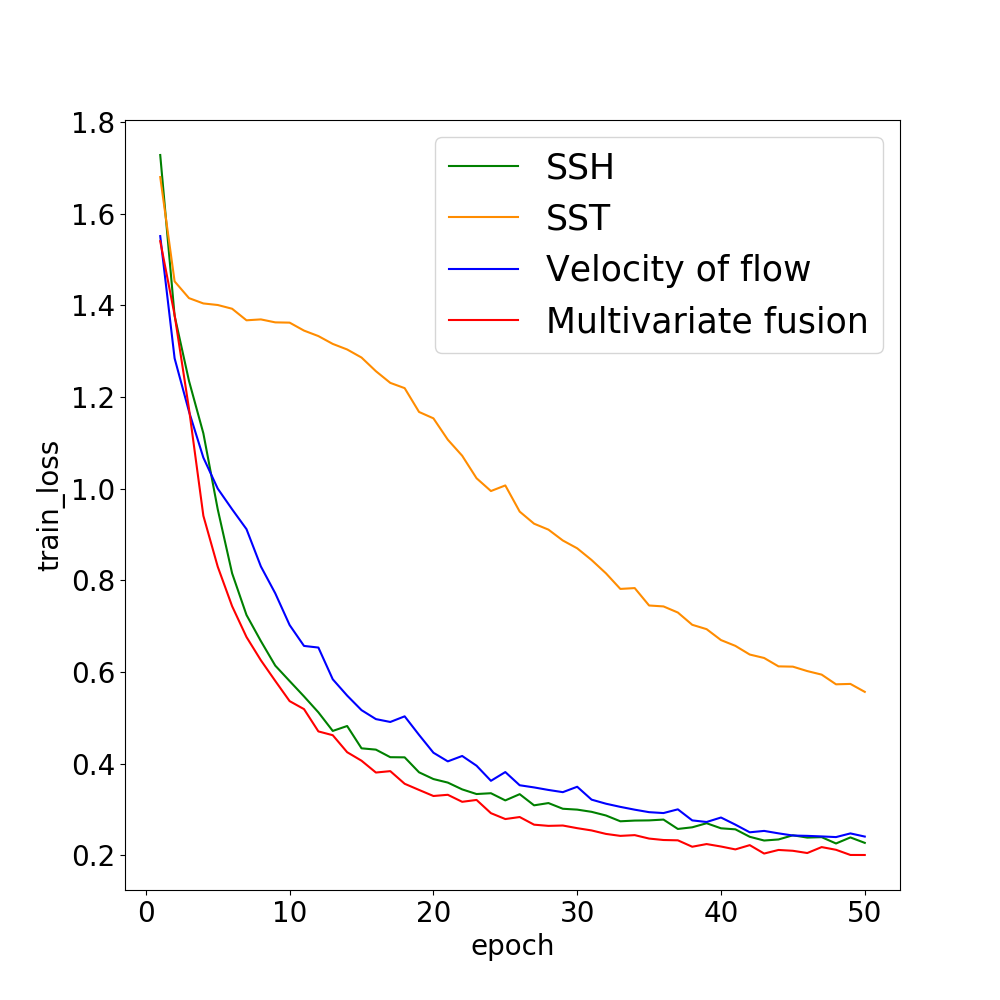}}
\subfigure[The accuracy curve]{
\includegraphics[width=5.5cm,height=5cm]{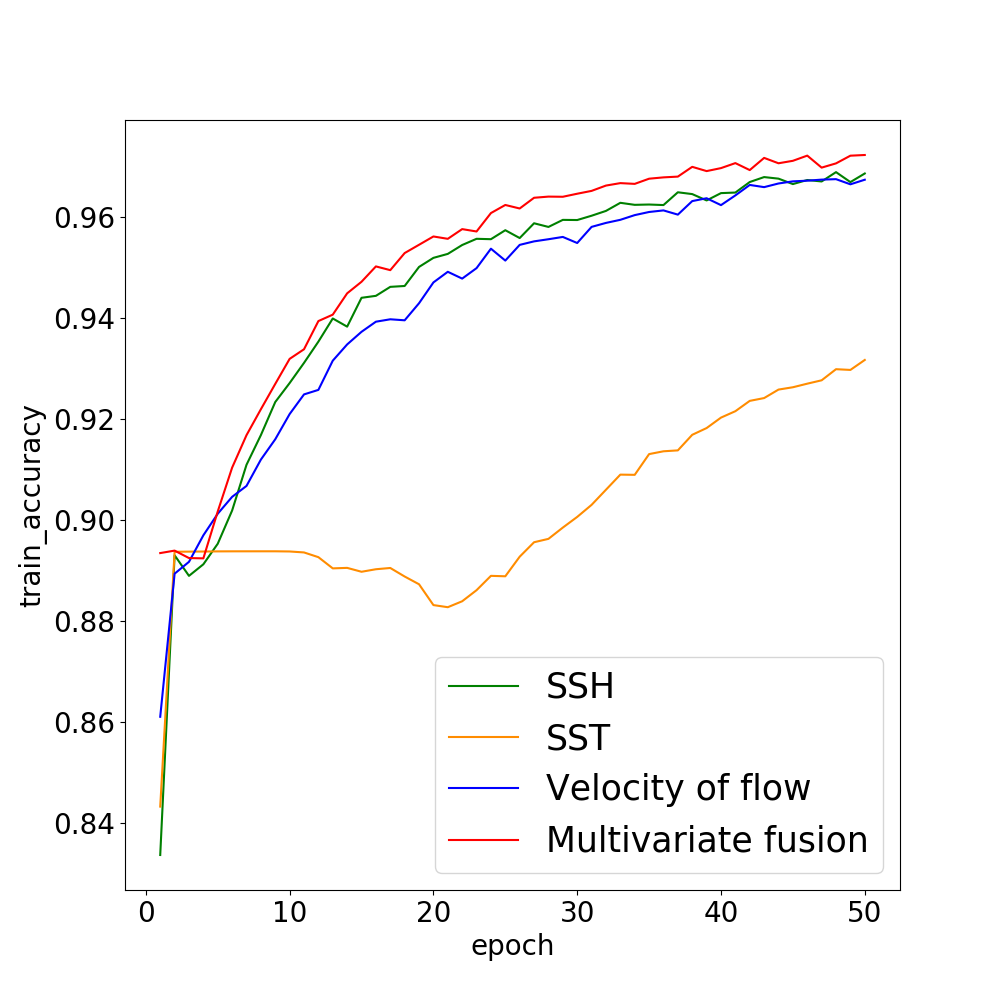}}
 \caption{The loss and accuracy curves of using our architecture on four training sets.}\label{fig5}
\end{figure}

To study the significance of our dataset that incorporates multiple variables, we trained our architecture on four datasets, which include SSH variable, SST variable, velocity of flow variable and multiple variables that combines these three variables respectively. Figure~\ref{fig5} shows the learning curves of using our architecture on four training sets, the green, orange, blue and red curve represents the learning curve of SSH, SST, velocity of flow and multivariate fusion training set separately. We can see from Figure~\ref{fig5} (a) that the loss is gradually decreasing and the loss on the multivariate training set is much lower than the loss on other training sets as epoch increases. Similarly, we can see that the accuracy is gradually increasing and the accuracy on the multivariate training set is much higher than the accuracy on other training sets as epoch increases in Figure~\ref{fig5} (b). By the way, we can also see the influence of the other three variables on mesoscale eddy detection from Figure~\ref{fig5}. There is no doubt that SSH is the most important factor among these three variables for mesoscale eddy detection, which is used as input variable in many papers. Next, velocity of flow is also relatively important for mesoscale eddy detection because it occupies two of the four channels. In comparison, the effect of SST on mesoscale eddy detection is not as strong as the former variables.

\begin{table}[ht]
\centering
\caption{The loss and accuracy of using our architecture on four training sets.}\label{t1}
\begin{tabular}{p{3cm}p{1.5cm}<{\centering}p{2cm}<{\centering}p{1.5cm}<{\centering}p{1.5cm}<{\centering}}
  \hline
\textbf {Dataset}          & Loss        & Crossentropy loss      & Dice loss      & Accuracy\\ \hline
SSH                          &	0.2351     &  0.0935                &  0.1314       &  96.77\%\\
SST                          &	0.5565     &  0.2144                &  0.2889       &  93.16\%\\
Velocity of flow                    &   0.2381    &  0.0940                &  0.1341       &  96.50\%\\
Multivariate fusion           & \textbf{0.1902}       & \textbf{0.0763\%}       & \textbf{0.1076\%}      &\textbf{97.32\%}\\
\hline
\end{tabular} 
\end{table}

Moreover, the loss and accuracy of using our architecture on these four training sets after 50 epochs are shown in Table~\ref{t1}, where the cross-entropy loss and dice loss of our loss are shown respectively.

\begin{table}[ht]
\centering
\caption{The accuracy of using our architecture on four testing sets.}\label{t2}
\begin{tabular}{p{4cm}p{4cm}<{\centering}}
  \hline
\textbf {Dataset}             &  Accuracy\\ \hline
SSH                              & 	96.84\%\\
SST                              &	91.64\%\\
Velocity of flow                        &  96.50\%\\
Multivariate fusion                &\textbf{97.06\%}\\
\hline
\end{tabular} 
\end{table}

Table~\ref{t2} shows the accuracy of using our architecture on these four testing sets to more convincingly prove that the results on our dataset have improved significantly. Through these comparative experiments on training sets and test sets, we can clearly see that the method based on multivariate fusion dataset surpasses the methods based on the other three datasets.

\begin{figure}[ht]
\centering
\subfigure[SSH visualization]{
\includegraphics[width=3cm,height=3cm]{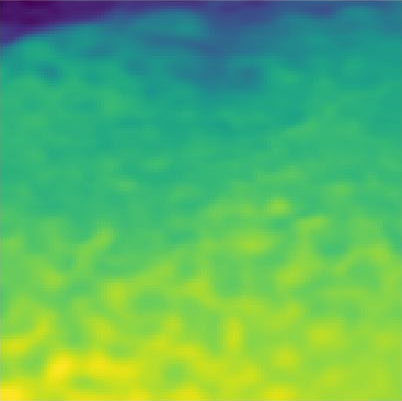}}
\subfigure[SSH]{
\includegraphics[width=3cm,height=3cm]{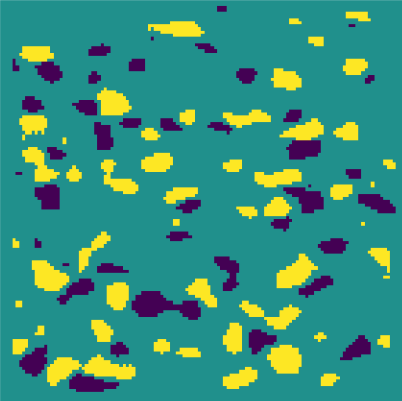}}
\subfigure[SST]{
\includegraphics[width=3cm,height=3cm]{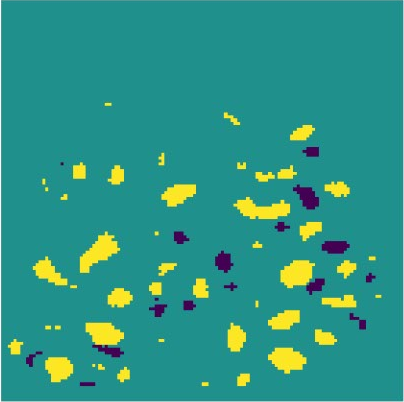}}
\subfigure[Velocity]{
\includegraphics[width=3cm,height=3cm]{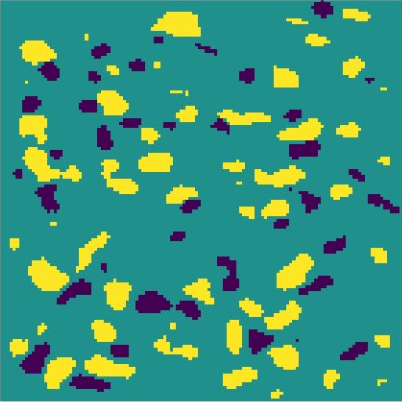}}
\subfigure[Multivariate fusion]{
\includegraphics[width=3cm,height=3cm]{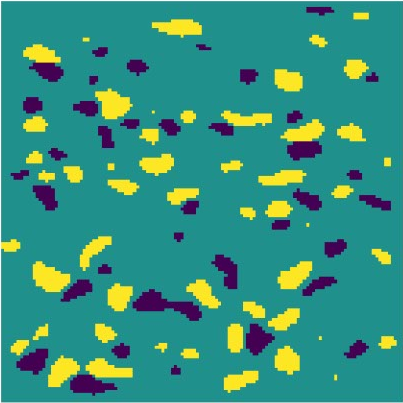}}
\subfigure[Groundtruth]{
\includegraphics[width=3cm,height=3cm]{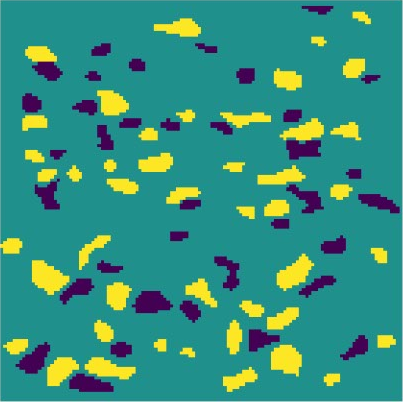}}
 \caption{The eddy segmentation results of using our architecture on four testing sets.}\label{fig6}
\end{figure}

In order to visually and intuitively demonstrate the improvement on our multivariate fusion dataset compared with the other three datasets, the examples of eddy segmentation results from the same region of sea using these four datasets can be seen in Figure~\ref{fig6}. Comparing with the groundtruth, we can see that the segmentation results using SST dataset misses a lot eddies and the segmentation result using velocity of flow dataset detects some extra eddies incorrectly. Although the segmentation based on SSH and our multivariate fusion dataset are similar, you will find that the segmentation based on our dataset is more accurate than the segmentation based on SSH dataset through watching the detail of segmentation carefully.


\subsubsection{Results on different architectures}

\begin{figure}[ht]
\centering
\subfigure[The loss curve]{
\includegraphics[width=5.5cm,height=5cm]{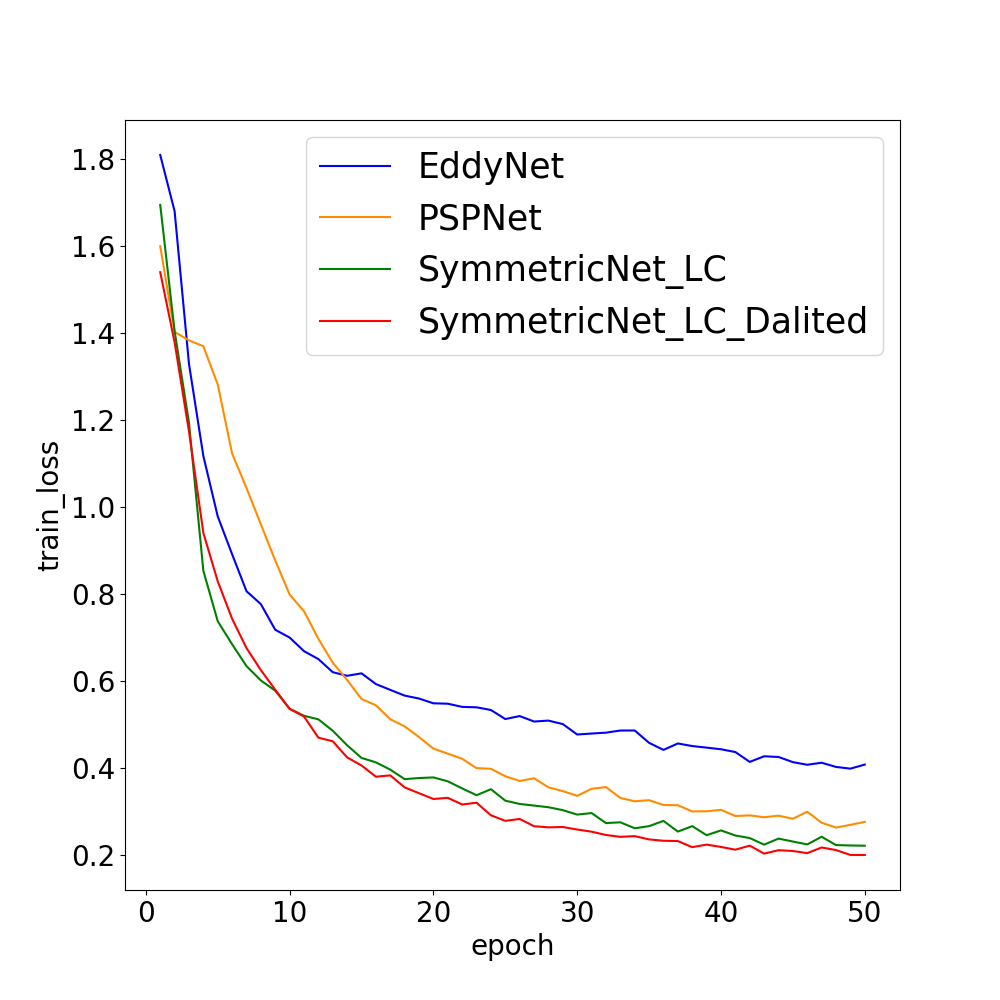}}
\subfigure[The accuracy curve]{
\includegraphics[width=5.5cm,height=5cm]{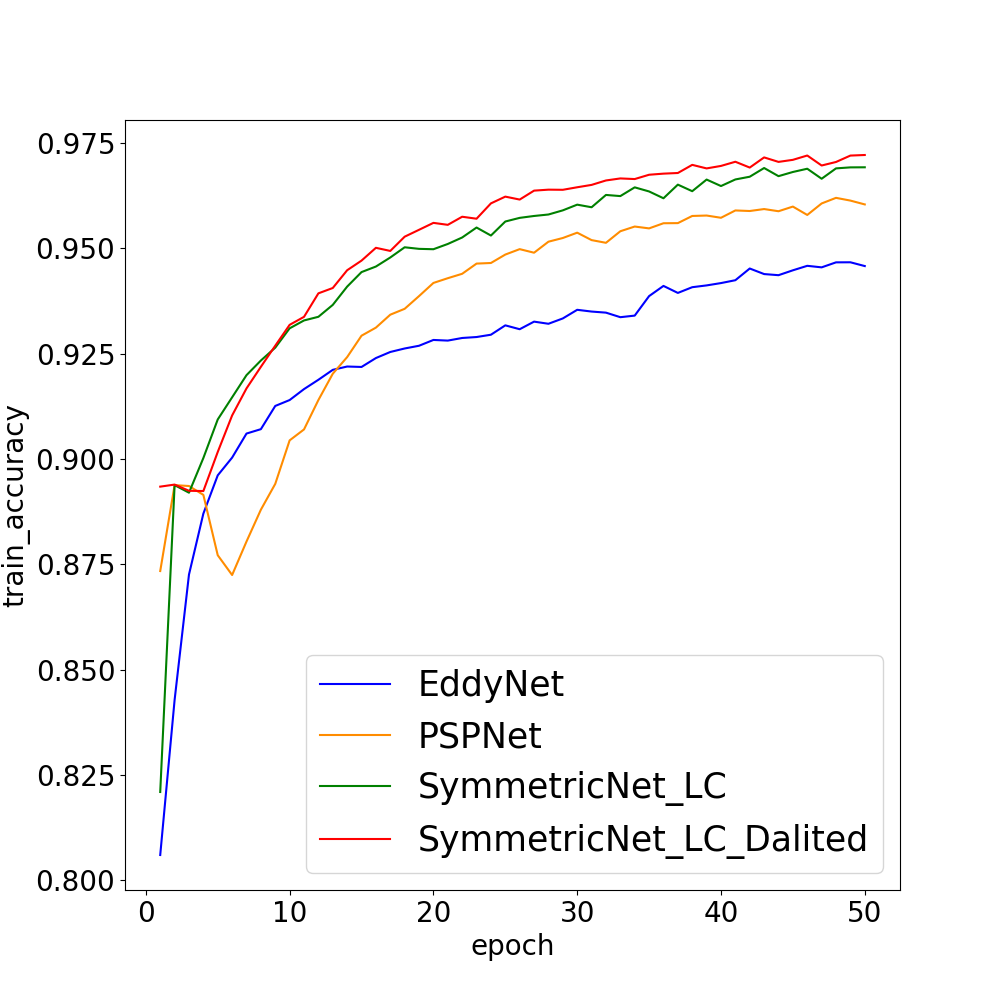}}
 \caption{The loss and accuracy curves of using different networks on multivariate fusion training set.}\label{fig7}
\end{figure}

The result on our architecture is compared with other architecture for verifying the impact of our framework. However, there are few mesoscale eddy detection methods based on deep learning. So we only choose EddyNet and PSPNet, which also use the idea of segmentation to deal with mesoscale eddy detection problem. Beside that, SymmetricNet without dilated convolution are the remaining comparison networks to prove the impact of our network(SymmetricNet with lateral connection and dilated convolution). It is to be noted that these networks is trained on our multivariate fusion dataset because we have proved the superiority of our dataset in Sec. 4.3.1. Figure~\ref{fig7} shows the learning curves of using different networks on multivariate fusion training set, the blue, orange, green and red curve represents the learning curve of EddyNet, PSPNet, SymmetricNet without dilated convolution and our network separately. Figure~\ref{fig7} (a) depicts that the loss is gradually decreasing and the loss of our architecture is much lower than the loss on other networks as epoch increases. Similarly, we also can see that the accuracy is gradually increasing and the accuracy of our network is much higher than the accuracy on other networks as epoch increases in Figure~\ref{fig7} (b).

\begin{table}[ht]
\centering
\caption{The loss and accuracy of using different networks on multivariate fusion training set.}\label{t3}
\begin{tabular}{p{2cm}|p{1cm}<{\centering}p{1cm}<{\centering}|p{1cm}<{\centering}p{2cm}<{\centering}p{1.5cm}<{\centering}p{1cm}<{\centering}}
  \hline
\textbf {Method}      & LC          & Dilated conv        & Loss        & Crossentropy loss      & Dice loss     & Accuracy   \\ \hline
Eddynet                 & --	      &	--                  &	0.4083    &  0.1636                &  0.2168       & 	94.58\%\\
PSPNet                  & --          &	--                  &	0.2766    &  0.1089                &  0.1542       &	96.05\%     \\
\hline
SymmetricNet            & $\surd$     & --                  &	0.2126    &  0.0867                &  0.1182       &    97.04\%  \\
SymmetricNet            & $\surd$     & $\surd$             &	\textbf{0.1902}    &  \textbf{0.0763}                &  \textbf{0.1076}       &\textbf{97.32\%} \\
\hline
\end{tabular} 
\end{table}

\begin{table}[ht]
\centering
\caption{The accuracy of using different networks on multivariate fusion testing set.}\label{t4}
\begin{tabular}{p{3cm}|p{2cm}<{\centering}p{2cm}<{\centering}|p{2.5cm}<{\centering}}
  \hline
\textbf {Method}      & LC          & Dilated conv        & Accuracy   \\ \hline
Eddynet                 & --	      &	--                  & 	93.77\%\\
PSPNet                  & --          &	--                  &	96.25\%     \\
\hline
SymmetricNet            & $\surd$     & --                  &   96.72\%  \\
SymmetricNet            & $\surd$     & $\surd$             &\textbf{97.06\%} \\
\hline
\end{tabular} 
\end{table}

\begin{figure}[ht]
\centering
\subfigure[SSH visualization]{
\includegraphics[width=3cm,height=3cm]{16.png}}
\subfigure[Eddynet]{
\includegraphics[width=3cm,height=3cm]{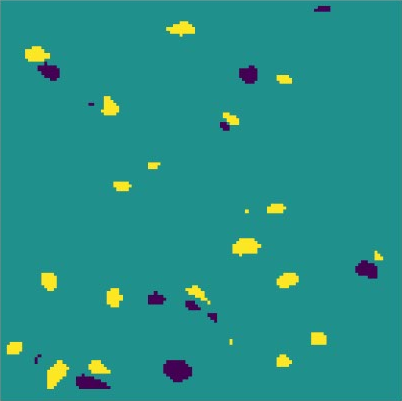}}
\subfigure[PSPNet]{
\includegraphics[width=3cm,height=3cm]{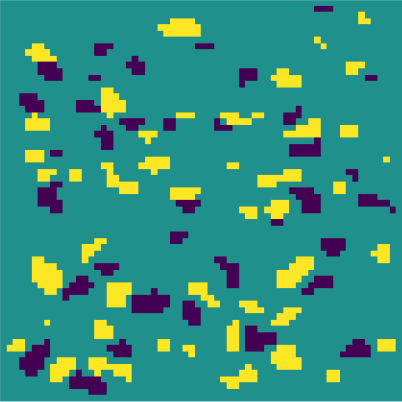}}
\subfigure[SymmetricNet(LC)]{
\includegraphics[width=3cm,height=3cm]{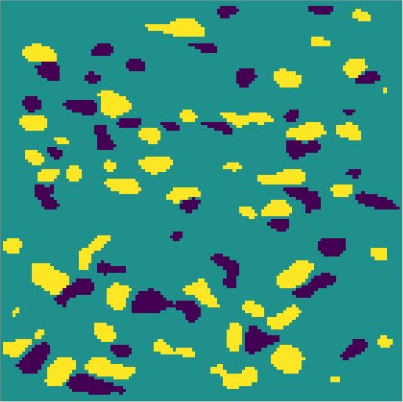}}
\subfigure[Our method]{
\includegraphics[width=3cm,height=3cm]{17.png}}
\subfigure[Groundtruth]{
\includegraphics[width=3cm,height=3cm]{18.png}}
 \caption{The eddy segmentation results of using different networks on multivariate fusion testing set.}\label{fig8}
\end{figure}

\begin{table}[ht]
\centering
\caption{The accuracy of using different networks on four different testing sets.}\label{t5}
\begin{tabular}{p{4.5cm}p{1.5cm}<{\centering}p{1.5cm}<{\centering}p{1.5cm}<{\centering}p{2cm}<{\centering}}
  \hline
\textbf {Method}      & SSH        & SST        & Velocity of flow   &  Multivariate fusion\\ \hline
Eddynet                 &   94.78\%  &	89.59\%   &	93.55\%          & 	93.77\%\\
PSPNet                 &	96.15\%  &	89.80\%   &	95.77\%          &	96.25\%     \\
\hline
SymmetricNet(LC)        &   96.37\%  &	89.83\%   &	95.90\%          &   96.72\%  \\
SymmetricNet(LC+Dilated)       &   96.84\%     &   91.64\%   &	96.50\%     &\textbf{97.06\%} \\
\hline
\end{tabular} 
\end{table}

\begin{figure}[ht]\centering

\subfigure[SSH visualization]{
\begin{minipage}[t]{0.25\linewidth}
\centering
\includegraphics[width=3cm,height=3cm]{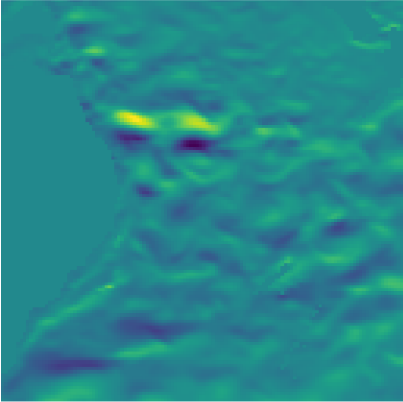}
\includegraphics[width=3cm,height=3cm]{16.png}
\end{minipage}%
}%
\subfigure[Our method]{
\begin{minipage}[t]{0.25\linewidth}
\centering
\includegraphics[width=3cm,height=3cm]{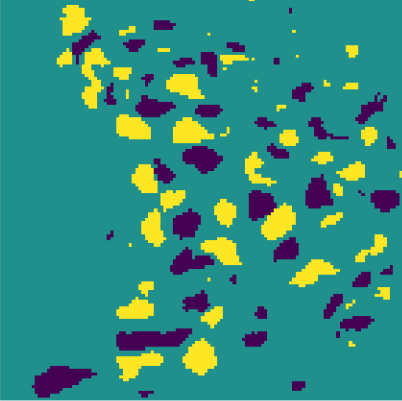}
\includegraphics[width=3cm,height=3cm]{17.png}
\end{minipage}%
}%
\subfigure[Groundtruth]{
\begin{minipage}[t]{0.25\linewidth}
\centering
\includegraphics[width=3cm,height=3cm]{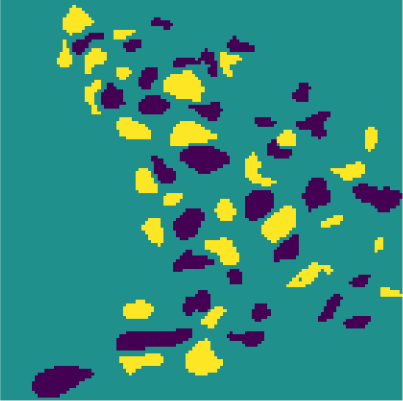}
\includegraphics[width=3cm,height=3cm]{18.png}
\end{minipage}
}%
\centering
\caption{The eddy segmentation results of our method.}\label{fig9}
\end{figure}

Same as the last section, we also give the loss and accuracy of using different networks on multivariate fusion training set after 50 epochs in Table~\ref{t3}, the accuracy of using different networks on multivariate fusion testing set in Table~\ref{t4}. We can found that our architecture yields 97.06\% in terms of accuracy, attaining 3.29 percentage points higher than EddyNet, 1.27 percentage points higher than PSPNet, and 0.4 percentage points higher than the SymmetricNet without dilated convolution our proposed. The examples of eddy segmentation  from the same region of sea using different networks on multivariate fusion testing sets is shown in Figure~\ref{fig8}. Apparently, the segmentation of our method is the most similar to groundtruth. However, the eddy segmentation results of EddyNet misses lots of eddies, PSPNet locates eddies inaccurately and the segmentation of SymmetricNet without only dilated convolution overestimates some eddies.

To demonstrate the best performance of using our architecture on our dataset more convincingly, the accuracy of using different networks on four different testing sets are shown in Table~\ref{t5}. We can find that our method is better than others no matter which dataset and our dataset is better than others no matter which network. In other words, our architecture on our dataset attain the best performance. Figure~\ref{fig9} shows the examples of eddy segmentation results using our method on our dataset, which have a resemblance to the groundtruth segmentation.

\subsubsection{Results on different loss functions}

In this work, we also put forward a novel loss function merging dice loss function with categorical cross-entropy loss function, which improves the detection performance. Dice loss function is popular in semantic segmentation, and categorical cross-entropy loss function has an extremely important influence in the classification problem. The comparison among our loss function against the crossentropy loss function and the dice loss function is shown in Table~\ref{t6} and Table~\ref{t7}. Regardless of loss or accuracy, our loss function achieves the best performance.

\begin{table}[ht]
\centering
\caption{The loss and accuracy of using different loss functions on training set.}\label{t6}
\begin{tabular}{p{3.5cm}p{2cm}<{\centering}p{1.5cm}<{\centering}p{1.5cm}<{\centering}p{1.5cm}<{\centering}}
  \hline
\textbf {Method}                  & Crossentropy loss     & Dice loss      & Our loss       & Accuracy \\ \hline
Only crossentropy loss              & 	0.0922              &	--            &	--             &	96.31\% \\
Only dice loss                      &	--                   &	0.1149       &	--             &	96.69\% \\
Our loss                            &   \textbf{0.0763}              &   \textbf{0.1076}       &	\textbf{0.1902}        &	\textbf{97.32\%} \\
\hline
\end{tabular} 
\end{table}

\begin{table}[ht]
\centering
\caption{The accuracy of using different loss functions on testing set.}\label{t7}
\begin{tabular}{p{4cm}p{4cm}<{\centering}}
  \hline
\textbf {Method}                        & Accuracy \\ \hline
Only crossentropy loss                   &	94.50\% \\
Only dice loss                           &	95.60\% \\
Our loss                                     &	\textbf{97.06\%} \\
\hline
\end{tabular} 
\end{table}

\section{Conclusion}

In this paper, we construct a novel multivariate fusion dataset, which is composed of SSH, SST, velocity of flow. Additionally, the SymmetricNet for mesoscale eddy detection is proposed, which merges low-level feature maps from the downsampling pathway and high-level feature maps from the upsampling pathway by lateral connection. In addition, the dilated convolutions are applied to reduce the loss of spatial information of the feature map caused by common convolutions. To evaluate our dataset and method, we make comparative experiment on different datasets, different architectures and different loss functions. In the consequence, our method based on our dataset attains striking detection performance. The experiment results also illustrates that our method on our dataset is better than the other existing comparable methods on different datasets and provides a simple but solid baseline.


\bibliographystyle{elsarticle-num}
\bibliography{mybibfile}

\end{document}